\documentclass[a4paper]{article}
\usepackage{tabularx}
\usepackage{INTERSPEECH2021}
\def\etal{\emph{et al.}}
\def\ie{\emph{i.e.}}
\def\eg{\emph{e.g.}}
\usepackage[hyphens]{url}

\title{Configurable Privacy-Preserving Automatic Speech Recognition}
\name{Ranya Aloufi, Hamed Haddadi, David Boyle}
\address{Imperial College London}
\email{\{ra6018,h.haddadi,david.boyle\}@imperial.ac.uk}

\begin{document}

\maketitle
\begin{abstract}
Voice assistive technologies have given rise to far-reaching privacy and security concerns. In this paper we investigate whether modular automatic speech recognition (ASR) can improve privacy in voice assistive systems by combining independently trained separation, recognition, and discretization modules to design configurable privacy-preserving ASR systems. We evaluate privacy concerns and the effects of applying various state-of-the-art techniques at each stage of the system, and report results using task-specific metrics (\ie~WER, ABX, and accuracy). We show that overlapping speech inputs to ASR systems present further privacy concerns, and how these may be mitigated using speech separation and optimization techniques. Our discretization module is shown to minimize paralinguistics privacy leakage from ASR acoustic models to levels commensurate with random guessing. We show that voice privacy can be \emph{configurable}, and argue this presents new opportunities for privacy-preserving applications incorporating ASR.  
  
\end{abstract}

\section{Introduction}
Voice assistive technologies 
facilitate seamless interactions between consumers and online services using Automatic Speech Recognition (ASR) technology. 
Deep learning has been a driving force in research and practice across speech application domains, yielding rapid advancements in ASR. 
Although deep models have comparable performance with more conventional approaches like hidden Markov model (HMM)-based ASR, considerable privacy vulnerabilities arise due to training models on real voice data that contain a significant amount of sensitive information~\cite{schuller1988emotion-COP}.

\begin{figure}[t!]
  \centering
  \includegraphics[width=\columnwidth]{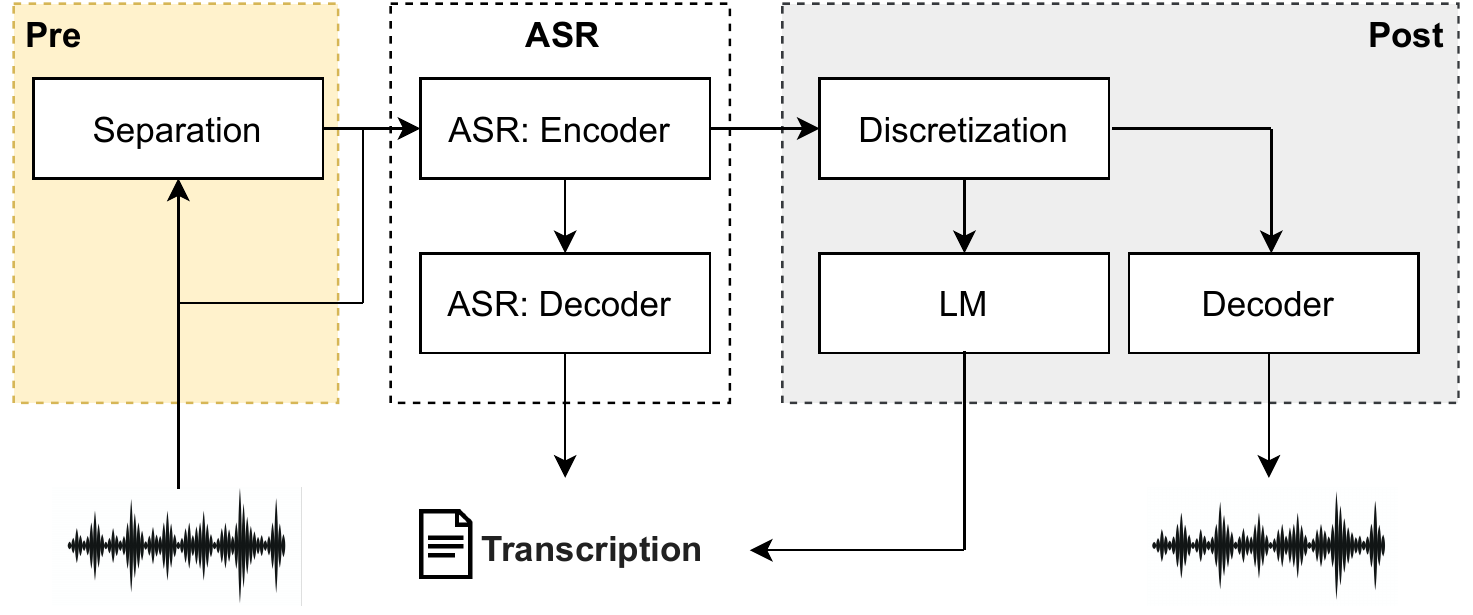}
  \caption{Configurable privacy-preserving ASR architecture}
      \label{fig:Modular}
\end{figure}

In this paper, we propose a modular system for privacy-preserving processing of voice data that combines independently trained separation, ASR, and discretization modules (Fig.~\ref{fig:Modular}). We study the effect of different state-of-the-art models at each stage of the pipeline and report results using task-specific metrics including word error rate (WER), zero-shot linguistics metrics~\cite{nguyen2020zero-COP} and downstream classification accuracy. The architecture must be modular to enable different privacy settings, thus we investigate the use of speech separation and speech discretization in an attempt to enhance privacy protection. Because the modules interact only at the I/O level, they can be implemented using different tools and independently optimized. Specifically, an obtained voice signal can be processed either by the Separation module to obtain a filtered signal belonging to a speaker of interest, or it may progress directly to ASR. The output of the separation module is fed to the ASR module. ASR systems consist of an encoder and a decoder to extract textual content from the voice signal. The output from the ASR module can be transferred to the Discretization module, which uses the output from the encoder of the ASR to extract the phonemic units (\ie~discrete, low bitrate, or pseudo-text units). Learning these discrete units can be used to extract a transcription directly by training a language model over it, or to obtain a new voice signal using a decoder trained on a system speaker (\ie~speech re-synthesis)~\cite{lakhotia2021generative-COP}. 

The \textbf{main contribution} of this paper is a configurable framework that enhances privacy in ASR systems based on speech separation and a discrete modular approach\footnote{Models and code will be publicly released on acceptance~\url{https://github.com/RanyaJumah/EDGY}}. A new benchmark is defined against which to compare possible privacy settings. We also introduce phoneticsMix, a new speech dataset derived from Libri-light and WHAM~\cite{Wichern2019WHAM-COP} noise to evaluate ASR (\ie~ABX) in mixed utterance and zero-resource settings. We experimentally evaluate the proposed approach and perform a systematic performance analysis for task-specific metrics. We show that Separation improves WER by $\sim$16\% in multi-speaker settings, and Discretization reduces classification accuracy in inferring sensitive attributes to 11\%-51\% (\ie~over multi/binary attributes).


\section{Voice Privacy}
\label{sec:related_work}

\subsection{Privacy-preserving Voice Analytics}
Privacy enhancing technology for voice assistive systems is an emerging and important area of research. Aloufi~\etal~describe scenarios whereby attackers can infer a significant amount of private information by observing the output of state-of-art deep acoustic models for speech processing tasks~\cite{aloufi2020privacypreserving-COP}. Nautsch~\etal,~in~\cite{nautsch-2019-COP}, demonstrate the importance of proposing and developing new privacy-preserving technologies for protecting speakers and speech characterization in voice signals. 

The recent VoicePrivacy initiative~\cite{tomashenko2020introducing-COP} promotes the development of anonymization methods that aim to suppress personally identifiable information in speech (\ie~speaker identity) while leaving linguistic content intact. Most of the proposed works focus on protecting/anonymizing speaker identity using voice conversion (VC) mechanisms~\cite{Hidebehind_2018-COP, srivastava2020design-COP, ahmed2020preech-COP}. These VC methods, however, aim to protect the speaker identity against different linkage attacks limited by the attacker’s knowledge~\cite{9053868-COP}. It has been found that when the attacker has complete knowledge of the VC scheme and target speaker mapping, none of the existing VC methods can protect speaker identity. Beyond speaker identity, various works propose to protect speaker gender~\cite{jaiswal2019privacy-COP} and emotion~\cite{emotionless_2019-COP}, wherein an edge-based system is proposed to filter affect patterns from a user's voice before sharing it with cloud services for further analysis.

Another direction is protecting users' privacy by ensuring that sensitive data is not unnecessarily transmitted to service providers. This may be done by optimizing the neural network architecture using quantization and pruning techniques to enable on-device processing. He~\etal, for example, propose an end-to-end speech recognizer for on-device applications such as voice commands and voice search that runs comfortably in real time on a Google Pixel phone~\cite{he2019streaming-COP}. Similarly, Aloufi~\etal~analyze optimization techniques including filter-pruning, weight-pruning, and quantization to learn privacy-preserving representations from raw data in near real-time with minimal model performance loss~\cite{aloufi2020paralinguistic-COP}. 
\subsection{Potential Threat}
Voice-controlled technologies that allow users to speak to interact with their devices are based on accurate speech recognition to ensure appropriate responsiveness. In many real-world usage scenarios, voice inputs may often consist of overlapping speech. This raises privacy concerns owing to the capture and processing of audio that may involve two or more people, and without their explicit consent. This violates GDPR provisions, for instance. Furthermore, the deep acoustics models used to analyze these recordings may encode more information than needed for the task of interest (\ie~ASR), such as profiles of users' demographic categories, personal preferences, emotional states, etc., and may therefore significantly compromise their privacy.

Protecting privacy requires more than hiding speaker information or running on-device ASR. We consider that privacy is subjective, with varying attitudes between users, which may even depend on the services (and/or service providers) with which these systems communicate. We therefore advocate the principle of \textit{configurable privacy}. Unlike other approaches, we seek to protect the privacy of multiple user attributes for various scenarios that depend on voice input or speech analysis, \ie~sanitizing the speech signal of attributes a user may not wish to share, but without adversely affecting functionality. We also emphasize the importance of enabling different privacy settings for optimizing the privacy-utility trade-off and promoting transparent privacy management practises.
\section{Configurable Privacy ASR}
\label{sec:system_overview}

\subsection{Speech Separation \& Discretization}

Personal voice activity detection (VAD)~\cite{ding2020personal-COP} systems detect the voice activity of a target speaker at the frame level, only triggering for the target user. This helps reduce computational cost and energy consumption, particularly in scenarios where a keyword detector is not preferable. VoiceFilter-Lite~\cite{wang2020voicefilterlite-COP} is a single-channel source separation model that runs on-device to preserve only the speech signals from a target user as part of a streaming speech recognition system. Similarly, Xue ~\etal~in~\cite{xue2020online-COP} propose a method called speaker tracing buffer, which can track speaker information consistently across the chunk by extending a self-attention mechanism to maintain the speaker permutation information determined in previous chunks. Target-speaker voice activity detection and E2E speaker-attributed ASR can thus play a significant role in improving E2E ASR performance on overlapped speech, as well mitigating privacy concerns.

Discovering discrete units may be useful for phonetic learning for low resourced languages or for languages with no textual resources, which cannot be addressed in the supervised setting, making speech technology more inclusive. One motivation behind discretization is that it can capture high-level semantic content from the speech signal \eg~phoneme due to the discrete nature of phonetic units~\cite{nguyen2020zero-COP}. Such representations could be used in downstream speech applications requiring symbolic or sparse input, \eg~for faster retrieval in speech search systems~\cite{peng2020correspondence-COP} or to build a synthesizer in a target speaker’s voice using the discovered units~\cite{kobayashi2021crank-COP}. 

\subsection{Design Overview}
Our pipeline consists of three basic components: Separation, ASR, and Discretization, as shown in Figure~\ref{fig:Modular}.

\begin{table*}[]
\caption{Overall performance of zero resource ASR on dev and other sets on ABX zero-shot metric (lower ABX is better)}
\label{tab:rl}
\begin{tabularx}{\textwidth}{ccccccccc}
\hline
\multicolumn{1}{l}{} & \multicolumn{4}{c}{ABX within} & \multicolumn{4}{c}{ABX across} \\ \hline
Model & \multicolumn{1}{l}{Dev-clean} & \multicolumn{1}{l}{Dev-O} & \multicolumn{1}{l}{DevM-clean} & \multicolumn{1}{l}{DevM-O} & \multicolumn{1}{l}{Dev-clean} & \multicolumn{1}{l}{Dev-O} & \multicolumn{1}{l}{DevM-clean} & \multicolumn{1}{l}{DevM-O} \\ \hline
CPC+DIS & 10.26 & 14.24 & 14.11 & 16.24 & 14.17 & 21.26 & 20.37 & 24.06 \\
SEP+CPC+DIS & 06.38 & 08.22 & 13.19 & 14.57 & 10.22 & 14.84 & 18.60 & 21.11 \\
wav2vec2+DIS & 16.47 & 24.30 & 28.41 & 30.78 & 17.41 & 27.43 & 20.86 & 29.28 \\
SEP+wav2vec2+DIS & 14.09 & 20.28 & 20.21 & 27.46 & 19.51 & 27.26 & 20.71 & 29.46 \\ \hline
\end{tabularx}
\end{table*}


\begin{table}[]
\caption{Overall performance of E2E ASR models on LibriCSS using WER metric (0S and 0L: 0\% overlap with short and long inter-utterance silence respectively)}
\label{tab:e2e}
\begin{tabular}{cllllll}
\hline
\multicolumn{1}{l}{} & \multicolumn{6}{c}{Overlap ratio in \%} \\ \hline
Method & \multicolumn{1}{c}{0L} & \multicolumn{1}{c}{0S} & \multicolumn{1}{c}{10} & \multicolumn{1}{c}{20} & \multicolumn{1}{c}{30} & \multicolumn{1}{c}{40} \\ \hline
CRDNN & 13.9 & 18.8 & 25.9 & 31.4 & 38.3 & 43.9 \\
SEP+CRDNN & 11.4 & 15.3 & 21.7 & 27.3 & 34.4 & 40.4 \\
TRANS & 14.4 & 16.4 & 23.5 & 33.0 & 35.3 & 45.2 \\
SEP + TRANS & 8.2 & 14.5 & 16.5 & 19.2 & 26.6 & 29.4 \\ \hline
\end{tabular}
\vspace{-5mm}
\end{table}

\subsubsection{Data \& Models}
We evaluate the proposed approach using a number of real-world datasets that were recorded for various speech processing purposes. We use the dev and test sets of Libri-light~\cite{9052942-COP} to evaluate the quality of discovering the discrete units from the raw recordings. 
We use LibriCSS~\cite{chen2020continuous-COP} and phoneticsMix to investigate the influence of overlapping speech in the raw recordings on E2E ASR systems and the separation role to reduce it. 
Inspired by LibriMix~\cite{cosentino2020librimix-COP}, we introduce phoneticsMix to investigate the effect of overlap speech on learning discrete units. The phoneticsMix dataset is derived from Libri-light, which was introduced mainly for training speech recognition systems under limited or no supervision (\ie~providing phonetic sequences as ground-truth), which we mixed with WHAM~\cite{Wichern2019WHAM-COP} noise to investigate speech separation for zero resources ASR that discovers its own units from raw speech. We use additional sets for various paralinguistics purposes, including speaker recognition (VoxCeleb~\cite{Nagrani_2017-COP}), accents and gender (Common Voice~\cite{ardila2019common-COP}), and emotion recognition (CREMA-D~\cite{cao2014crema-COP} and SAVEE~\cite{jackson2014surrey-COP}) to estimate the privacy leakage might cause by deep acoustics model of ASR systems.

We conduct our experiments using various state-of-the-art (SOTA, Table 3) models per module in the proposed framework. We first use SepFormer~\cite{subakan2021attention-COP} as a speech separation model which is an RNN-free architecture that employs a masking network composed of transformers only. We then implement our ASR experiments on CRDNN with connectionist temporal classification (CTC)/Attention-based and a Transformer-based E2E ASR model trained on Librispeech within SpeechBrain~\cite{SB2021-COP}. To enable speech discretization, we follow Zero Resource 2021 baseline~\cite{nguyen2020zero-COP} by concatenating self-supervised representation learning encoder with k-means clustering to convert continuous representations into discrete representations. We use the pre-trained contrast predictive coding (CPC)~\cite{nguyen2020zero-COP} and wav2vec 2.0~\cite{baevski2020wav2vec-COP} to encode the raw recordings into continuous representations and then train k-means on top of these representations.
\subsubsection{Speech Separation}
\label{sec:ss}

\textbf{Q1:}~\emph{a.) How does overlapping speech affect users' privacy? b.) Can speech separation help enhance ASR performance and accurately identify the speaker of interest to promote further privacy-preserving applications?}

Most prior studies concerning voice signal privacy have been constrained to mixtures of clean, near anechoic speech, not representative of many real-world scenarios. Overlapping speech influences a significant part of conversational interactions, and this may pose challenges for many speech technologies including E2E ASR and speaker verification. This is because they usually assume one or zero speakers to be active at the same time. Speech separation might therefore provide a solution for this overlapping speech problem. It might also lead to robust performance for voice assistive systems and a potential solution to improve privacy. 
We are interested in a scenario where multiple speakers may be present during raw recording. We first evaluate the performance of state-of-the-art ASR models in transcribing mixed utterances. We assess the influence of mixed utterances in learning speech representations from the raw signals. We then apply a single-channel source speech separation model to extract the speech signals from a target user only (\ie~use only the first channel data (monaural audio) for all our experiments), assuming this single channel belongs to that user and estimate the effect of this module in enhancing the performance in waveform or representations levels.

\textbf{Metric.}
We report the word error rate (WER) on LibriCSS~\cite{chen2020continuous-COP} evaluation set to estimate the performance of the E2E ASR models (\ie~using waveform). For representations level, we run ABX metric~\cite{nguyen2020zero-COP} on the Libri-light dev and test sets to estimate the discriminability between phonemes (lower is better). ABX is calculated by computing the distance between the representations associated with three acoustic tokens (\emph{a}, \emph{b}, and \emph{x}), two of which belong to the same category \emph{A} (\emph{a} and \emph{x}) and one which belongs to a different category \emph{B} (\emph{b}). Thus, the score is the estimated probability that \emph{a} and \emph{x} are closer to one another than \emph{a} and \emph{b}. We compute it within speaker (in which case, all stimuli \emph{a}, \emph{b}, and \emph{x} are uttered by the same speaker) and across speakers (\emph{a} and \emph{b} are from the same speaker and \emph{x} from a different speaker).

\textbf{Results.} We conduct two types of experiments to measure the performance of the learning task, E2E ASR and zero-resource ASR, using clean and mixed utterances to simulate overlapped interactions in realistic voice assistive deployment environments.\\

\textbf{No-SEP.} No separation mechanism is applied. As shown in Table~\ref{tab:rl}, the overlapped speech critically degraded the ASR performance in both settings (\ie~E2E and zero-resource) about 8\%-40\% WER and 6\%-8\% for E2E and zero-resource ASR, respectively. While we test it over mixed utterances (\ie~multi-speaker) only, this decrease may be worse with a noisier and more overlapped percentage. 

\textbf{With-SEP.} Even a small amount of overlap (10\%) critically degraded the ASR performance, however, the single-channel separation model might help improve its performance. The result is much better than the “No-SEP” scenario, showing that the single-channel separation model can improve the WER accuracy by 2\%-16\% in E2E ASR and about 2\% ABX score in zero-resource ASR (see Table~\ref{tab:rl}). The latter, however, needs more investigation to understand the effect of the mixed environments on the phonetic level, and more separation mechanisms may be needed for further improvement since zero-resource ASR developed under an unsupervised manner.
Once we tune this separation stage, the output can be connected to various applications, which could be speech recognition or speaker verification.
\begin{table*}[]
\caption{Performance for paralinguistic tasks (*Results incorporate a subset of Common Voice for accent/gender recognition)}
\label{tab:pp}
\centering
\begin{tabular}{clccccc}
\hline
\multicolumn{2}{c}{Task} & \begin{tabular}[c]{@{}c@{}}Speaker \\ Recognition\end{tabular} & \multicolumn{2}{c}{\begin{tabular}[c]{@{}c@{}}Emotion \\ Recognition\end{tabular}} & \begin{tabular}[c]{@{}c@{}}Accent \\ Identification\end{tabular} & \begin{tabular}[c]{@{}c@{}}Gender \\ Recognition\end{tabular} \\ \hline
\multicolumn{2}{c}{Model/Dataset} & VoxCeleb1 & CREMA-D & SAVEE & Common Voice* & Common Voice* \\ \hline
\multicolumn{2}{c}{SOTA} & 80.5 & 74.0 & 68.5 & - & - \\ \hline
\multicolumn{2}{c}{CRDNN+RLM} & 39.07 & 52.23 & 66.40 & 36.54 & 52.82 \\
\multicolumn{2}{c}{TRANS+TLM} & 47.04 & 59.31 & 71.53 & 38.56 & 57.51 \\
\multicolumn{2}{c}{CPC} & 45.11 & 62.02 & 57.91 & 71.69 & 82.65 \\
\multicolumn{2}{c}{wav2vec2} & 37.04 & 51.49 & 51.52 & 40.31 & 52.5 \\ \hline
\multicolumn{2}{l}{CRDNN+RLM+DIS} & 19.03 & 20.31 & 16.91 & 19.93 & 43.82 \\
\multicolumn{2}{c}{TRANS+TLM+DIS} & 20.14 & 23.35 & 20.83 & 20.68 & 44.21 \\
\multicolumn{2}{c}{CPC+DIS} & 17.04 & 29.19 & 19.33 & 23.72 & 44.51 \\
\multicolumn{2}{c}{wav2vec2+DIS} & 14.54 & 25.88 & 18.15 & 15.06 & 41.32 \\ \hline
\end{tabular}
\end{table*}

\subsubsection{Speech Discretization}
\textbf{Q2:}~\emph{a.) Can deep acoustic models of E2E ASR reveal sensitive attributes about users? b.) Can discrete units (e.g., phonemes) help to limit the amount of unnecessarily sensitive information learned by these models?}

Discrete units (\eg~phonemes) highlight linguistically relevant representations of the speech signal in a highly compact format~\cite{nguyen2020zero-COP} while being invariant to speaker-specific and background noise details. These additional acoustics information might lead to accurately inferring users' sensitive and private attributes (\eg~their gender, emotion, or health status). For example, an attacker (\eg~a `curious' service provider) may use this information encoded by speech recognition or speaker verification system to learn further sensitive attributes from user input with high inferring accuracy ranging from 40\% to 99.4\%, significantly better than guessing at random~\cite{aloufi2020privacypreserving-COP}. 

One motivation to apply discretization is that these discrete units can capture high-level semantic content from the speech signal, \eg~phoneme due to the discrete nature of phonetic units. We investigate the effectiveness of \emph{learning discrete units} in preserving the sensitive attributes in the encoded speech data. We extract an embedding using the encoder model of E2E ASR and self-supervised encoders (\ie~CPC and wav2vec2), and then discretize the output of these encoders using k-means. We use k-means as method to convert continuous frame representations into discrete representations using centroids of clusters obtained by training k-means on LibriSpeech. Each audio file will thus be discretized to a sequence of discrete units corresponding to the assigned clusters.

\vspace{-0.1mm}
\textbf{Metric.}
We estimate the level of the paralinguistics information preserved in encoded linguistics representations generated through ASR by measuring the classification accuracy of various paralinguistics tasks (\ie~speaker recognition, emotion recognition, accent identification, and gender recognition). Specifically, we test the accuracy in predicting these attributes over random guessing by training shallow models using Scikit-Learn on top of these representations.

\vspace{-0.1mm}
\textbf{Results.} We first evaluate the quality of the linguistics representations by aiming for high performance in the linguistics task (Section~\ref{sec:ss}), while achieving low performance on paralinguistics tasks (\ie~increase in inference accuracy over random guessing).

\textbf{No-DIS.} No discretization mechanism is applied. The accuracy of paralinguistics attributes in Table~\ref{tab:pp} (\ie~upper rows) shows that the possibility to infer speaker-sensitive information ranging from 36.5\%-82.7\% (\ie~for binary (gender) and multi-class (speaker, emotion, and accents) attributes), namely that a predictor of speaker characteristics can be learned from deep acoustics models used by ASR systems.

\textbf{With-DIS.}
As shown in Table~\ref{tab:pp}, the accuracy of the paralinguistics classification models dropped by 11\%-51\%, which is close to or less than for random guessing. For example, that the sensitive attribute in question is the speaker's `emotion', we have seven labeled categories in the available datasets (CREMA-D and SAVEE). The random guess rate for success is therefore around 14\%. If we assume that the sensitive attribute is `gender' (\eg~binary male or female), the random guess rate will be 50\%. Thus, such discretization may help to improve obtaining good language representations while reducing irrelevant paralinguistics information that may pose a threat to the user privacy.

Once we tune this discretization stage, the output can be connected to different downstream purposes, which could be direct transcription or speech resynthesis~\cite{lakhotia2021generative-COP}.
\section{Discussion}
\label{sec:discussion}

Most recently proposed solutions for privacy preservation fall between anonymization and cryptography~\cite{tomashenko2020introducing-COP}. Although these solutions were effective in anatomizing the speaker identity, they may prevent speaker authentication regarding voice assistive systems activation. These solutions may be useful from singular perspectives and for achieving a specific goals, like the identity of the speaker, but might fail to sufficiently address configurable privacy. Therefore, we highlight the need for a `configurable privacy' framework that can process the speech data in a privacy-preserving, task-related, and modular manner. 
First, we consider concern about overlapping speech on privacy, assuming the real-time deployment of voice assistive includes noisy and multi-speaker environments. Such environments could significantly affect the performance of ASR technology used by these assistive systems, leading to misactivations~\cite{PETS_2020smarthome-COP} or even open a door for further security attacks (\eg~adversarial examples~\cite{abdullah2020sok-COP}) to fool their prediction. We therefore show one potential mitigation by highlighting the importance of speech separation as a pre-step for voice assistive systems. As shown in Table~\ref{tab:e2e}, our experiments over mixed and noisy recordings show that these variations lead to a notable impact on E2E ASR, exhibiting $\sim$40\% drop in WER, whereas using speech separation can help to improve the performance by 16\%. In future work, we plan to extend the pre-step modules to include speaker diarization or even target speaker voice detection as a guide to the separation process and examine how that may improve privacy.

We also focus on the overlearning problem caused by deep acoustics models used for E2E ASR and speaker verification, revealing additional/sensitive information about the speaker from the raw recordings. The ability to discover the discrete units from raw signals and train language models (LMs) would go a long way towards making language technologies more inclusive, and encoding more acoustic information (\eg~speaker-specific) can be detrimental in achieving this purpose~\cite{nguyen2020zero-COP}. As shown in Table~\ref{tab:pp}, the inference accuracy significantly decreased after implementing discretization by about 11\%-51\%. Learning discrete units can also help to protect these sensitive attributes. In addition, transferring these units might be useful for compressing and transmitting voice signals~\cite{kleijn2021generative-COP}. Discretization, therefore, might improve voice communication in terms of privacy and processing efficiency, especially for languages with limited resources.

\section{Conclusions}
\label{sec:conclusion}

We presented a modular framework that includes three basic components: Separation, ASR, and Discretization, to reconcile speech processing technologies and privacy restrictions for voice assistive systems. We studied the performance of E2E ASR under variation in recording equipment (e.g. noisy conditions) and multi-speaker conversational scenarios and showed speech separation may facilitate privacy-preserving applications. We then investigated speech discretization to minimize the privacy leakages against inferences of users' sensitive or private attributes. Our experiments indicate that using speech separation can help to improve performance by 16\% and that discretization can effectively mitigate the overlearning issue caused by deep acoustics models. In the next steps of our work, we intend to extend our framework by consolidating recent developments with neural methods on speech processing technologies and progress towards more configurable, efficient, and privacy-preserving voice assistive technologies.


\bibliographystyle{IEEEtran}
\bibliography{mybib}


\end{document}